\documentclass[letterpaper, 10pt, conference]{ieeeconf}
\IEEEoverridecommandlockouts
\makeatother

\usepackage{cite}
\usepackage{amsmath,amssymb,amsfonts}
\usepackage{algorithmic}
\usepackage{graphicx}
\usepackage{textcomp}
\usepackage{amssymb}
\usepackage{multirow}
\usepackage[table]{xcolor}
\makeatletter
\let\NAT@parse\undefined
\makeatother
\usepackage[colorlinks,
            linkcolor=red,
            anchorcolor=blue,
            citecolor=green]{hyperref}
\def\BibTeX{{\rm B\kern-.05em{\sc i\kern-.025em b}\kern-.08em
    T\kern-.1667em\lower.7ex\hbox{E}\kern-.125emX}}

\begin{document}

\title{\LARGE \bf Large receptive field strategy and important feature extraction strategy in 3D object detection*}
\author{Leichao Cui$^{1}$, Xiuxian Li$^{1}$, Min Meng$^{1}$, and Guangyu Jia$^{2}$
\thanks{*This work was supported by the National Natural Science Foundation of China under Grant 62003243 and Grant 62103305, and the Shanghai Municipal Science and Technology Major Project, No. 2021SHZDZX0100.}
\thanks{$^{1}$Department of Control Science and Engineering, College of Electronics and Information Engineering, and the Shanghai Research Institute for Intelligent Autonomous Systems, Tongji University, Shanghai 201800, China. {\tt\small 2130715@tongji.edu.cn, 
        xxli@ieee.org,
        mengmin@tongji.edu.cn}}%
\thanks{$^{2}$the Centre for Robotics Research, Department of Engineering, King's College London, Strand, London WC2R 2LS, United Kingdom. {\tt\small guangyu.jia@kcl.ac.uk}}%
}

\maketitle

\begin{abstract}
The enhancement of 3D object detection is pivotal for precise environmental perception and improved task execution capabilities in autonomous driving. LiDAR point clouds, offering accurate depth information, serve as a crucial information for this purpose. Our study focuses on key challenges in 3D target detection.
To tackle the challenge of expanding the receptive field of a 3D convolutional kernel, we introduce the Dynamic Feature Fusion Module (DFFM). This module achieves adaptive expansion of the 3D convolutional kernel's receptive field, balancing the expansion with acceptable computational loads. This innovation reduces operations, expands the receptive field, and allows the model to dynamically adjust to different object requirements.
Simultaneously, we identify redundant information in 3D features. Employing the Feature Selection Module (FSM) quantitatively evaluates and eliminates non-important features, achieving the separation of output box fitting and feature extraction. This innovation enables the detector to focus on critical features, resulting in model compression, reduced computational burden, and minimized candidate frame interference.
Extensive experiments confirm that both DFFM and FSM not only enhance current benchmarks, particularly in small target detection, but also accelerate network performance. Importantly, these modules exhibit effective complementarity.
\end{abstract}

\section{INTRODUCTION}
Developing a comprehensive autonomous driving system that can be applied to diverse scenarios faces significant challenges. The perception system plays a crucial role in extracting and processing raw sensor data, serving as the initial module for information acquisition and processing. This information is then transmitted to regulation and control modules for further processing. The main goal of the perception system is to collect semantic and geometric data from the environment, such as vehicles and lane lines. Therefore, 3D object detection, at the heart of the perception system, plays a critical role and serves as the foundation for autonomous driving.\par
Point clouds obtained through LiDAR exhibit distinctive characteristics. In contrast to images arranged in a regular grid on a two-dimensional plane, point clouds manifest as a sparse and disorganized representation of three-dimensional (3D) data. Moreover, these point clouds provide precise depth information unaffected by external environmental conditions, such as variations in lighting, which can impact camera-based systems. Consequently, point clouds can faithfully portray the structure and contours of 3D objects. This inherent feature affords LiDAR a distinct advantage in 3D object detection.\par
\begin{figure}[t]
  \centering
  \includegraphics[width=0.43\textwidth]{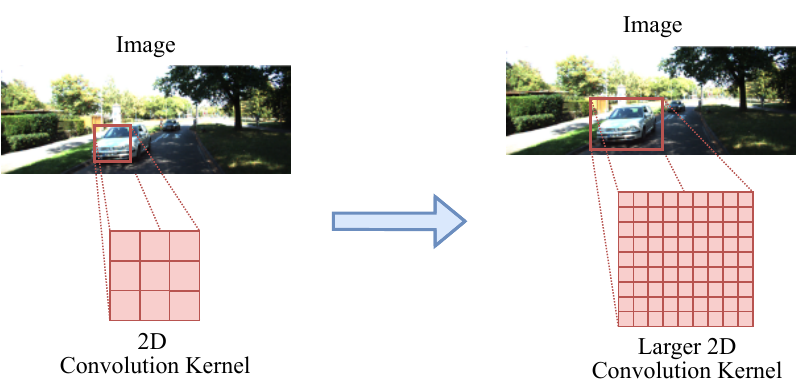}
  \vspace{-1em}
  \caption{The sizeable receptive field helps fully understand the object's overall structure and the surrounding environment's contextual information.}
  \vspace{-2em}
  \label{fig1.1}
\end{figure}
Hence, it is essential to improve the performance of LiDAR-based 3D detection networks in complex scenes, ensuring their robust detection capabilities and processing efficiency in diverse and challenging environments.
Despite major advances in current LiDAR-based 3D object detection methods, specific challenges remain and require solutions.\par
One concern pertains to extending the receptive field of the 3D convolutional kernel. Illustrated in Fig. \ref{fig1.1}, a substantial receptive field proves advantageous by encompassing a broader range of input data. This, in turn, facilitates the assimilation of more comprehensive global information and contextual features, allowing for a more profound modeling of features at an elevated level. Recent research endeavors\cite{SUY31,ConvNext} have identified that in 2D convolution-based networks, smaller convolution kernels are commonly employed for feature extraction. Consequently, attempts have been made to augment the receptive field by increasing the size of the convolution kernel, thereby enhancing the network's overall performance.\par
Nevertheless, 3D point clouds are not a simple extension of 2D images. The challenge arises as the enlargement of the convolution kernel size results in the cubic growth of computational and parametric quantities for 3D convolution. For instance, when transitioning from a 3$\times$3$\times$3 to a 9$\times$9$\times$9 3D convolution kernel, the increase in parameter count and computational load is much greater than that of a 2D convolution kernel operating under the same conditions.  Simultaneously, the sparsity inherent in point cloud data poses a limitation, as the data volume and processing capacity struggle to handle the rapidly increasing number of parameters. Consequently, the optimization of the network becomes imperative.\par
\begin{figure*}[thpb]
  \centering
  \includegraphics[width=0.9\textwidth]{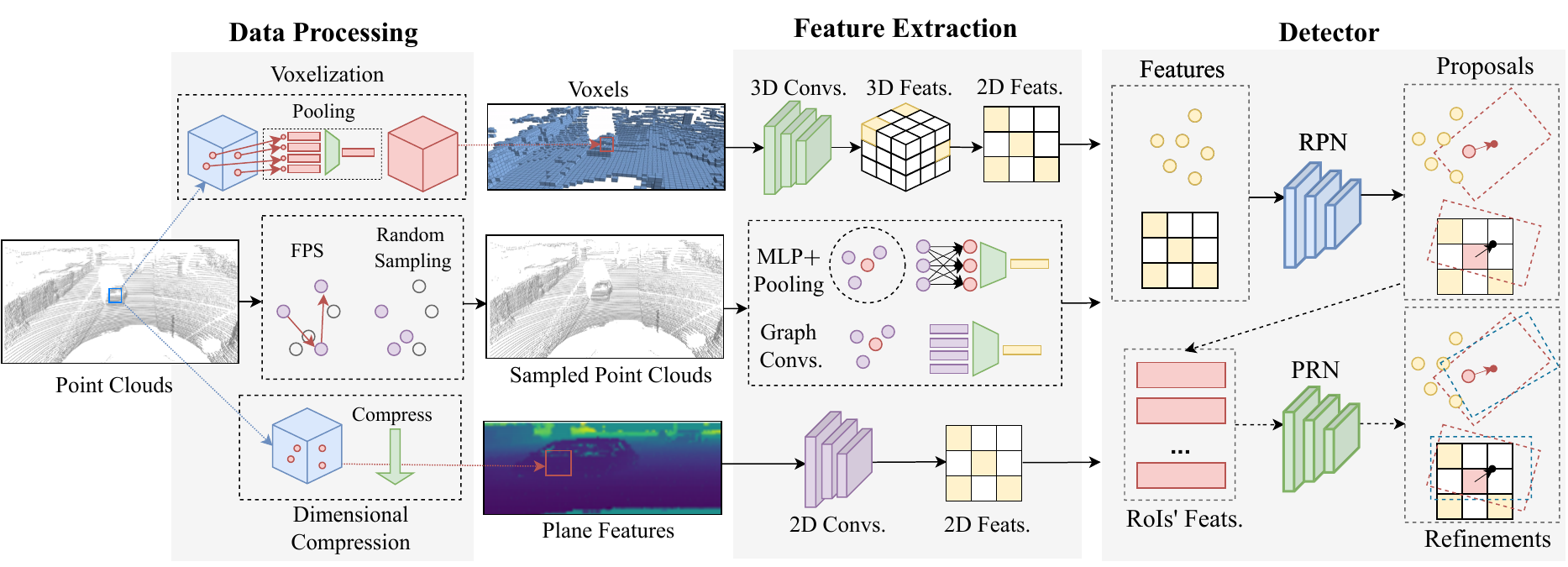}
  \vspace{-1em}
  \caption{The visual architecture of the point cloud detection network consists of three key components: data processing, feature extraction, and detector.  (\textbf{a}) Point clouds undergo various data processing methods for transformation. (\textbf{b}) Feature extraction employs diverse operations. (\textbf{c}) Extracted features are input into the detection network for object detection.}
  \vspace{-1.5em}

  \label{fig2.1}
\end{figure*}
Therefore, we introduce the Dynamic Feature Fusion Module (DFFM) to broaden the receptive field of the 3D convolutional kernel. DFFM dynamically extends the receptive field based on actual demand, ensuring that parameters remain within an acceptable range.\par
We additionally note that the features supplied to the 3D detector often include extraneous information. This causes the detector to execute numerous invalid computations, inefficiently utilizing valuable computational resources. Moreover, this feature disparity may lead to the hiding or masking of information with genuine value, posing challenges for the detector in learning them.
At the same time, this can also trigger mutual interference between multiple candidate boxes because the Non-Maximum Suppression (NMS) operation in the post-processing stage is exclusive, which eliminates the detection results with high overlap, thus possibly removing some valid targets and reducing the overall performance of the detector.\par
Thus we devise a plug-and-play Feature Selection Module (FSM) to effectively filter features, enabling the detector to concentrate on high-quality feature regression. The module decouples the output box fitting and feature selection, seamlessly integrating into existing network architectures without necessitating intricate modifications to the detector's design.\par
To validate the effectiveness of the above method, we improve on the existing 3D object detection networks\cite{VoxelNext, Second}. Experiments on the KITTI dataset demonstrate that our approach significantly improves performance while maintaining a faster detection speed.\par
Our contributions can be summarized as follows:
\begin{itemize}
\item	Introduction of the DFFM to address the computational challenges associated with an expanding receptive field, enhancing overall model optimization.
\item	Proposal of a plug-and-play FSM designed to eliminate non-essential features. This enables the detector to concentrate on fitting crucial features.
\item Demonstration of our network's superior performance over existing benchmarks, showcasing improvements in detection on the KITTI dataset, particularly for small objects.
\end{itemize}

\section{RELATED WORKS}
\subsection{LiDAR-based 3D Object Detection}
\textbf{Visual architecture.} The visual-like architecture of the point cloud detection network, as depicted in Fig. \ref{fig2.1}, is structured into three primary components: data processing, feature extraction, and the detector. Initially, the data processing phase transforms point cloud data into various representations. Following this, the feature extraction network and 3D object detector are employed to extract features and predict 3D bounding boxes on the processed data. \par
Specifically, the initial point clouds undergo processing through farthest point sampling\cite{Pointnet++}, random downsampling\cite{ngiam}, F-FPS\cite{3DSSD}, and coordinate refinement\cite{Pointformer} to yield the processed point clouds. Voxelization\cite{Second, SASSD, CenterPoint, PartA2, PV-rcnn,MMFusion} is applied to obtain voxels, while compressive mapping\cite{VeloFCN, LMNet} generates 2D feature maps. Subsequently, the processed voxels, point clouds, and 2D feature maps undergo 3D sparse convolution\cite{Second}, graph convolution\cite{GraphNet} or Set Abstraction\cite{Pointnet++}, and 2D convolution\cite{Resnet, swintransformer}, respectively. Finally, the output is derived through a detector based on point clouds\cite{Pointrcnn, 3DSSD, CenterPoint} or feature maps\cite{Second, Pointpillars, PV-rcnn}.\par
\textbf{Transformer architecture.} In contrast to traditional visual architecture, the Transformer architecture places a greater emphasis on end-to-end unified forms of data processing using the attention mechanism.
PT\cite{PT} and PCT\cite{PCT} have devised attention modules tailored for point cloud structures, employing the self-attention mechanism within the local point set. Conversely, Voxel Transformers\cite{Voxel-transformer} leverage the attention mechanism instead of the convolutional feature extraction backbone to address the sparse nature of voxels. Additionally, a dilated attention mechanism is proposed to expedite computation. SWFormer\cite{Swformer} prioritizes voxel feature diffusion and multi-scale fusion through the utilization of the attention mechanism.

\subsection{Large Receptive Field Strategy}
\begin{figure*}[thpb]
      \centering
      \includegraphics[width=0.7\textwidth]{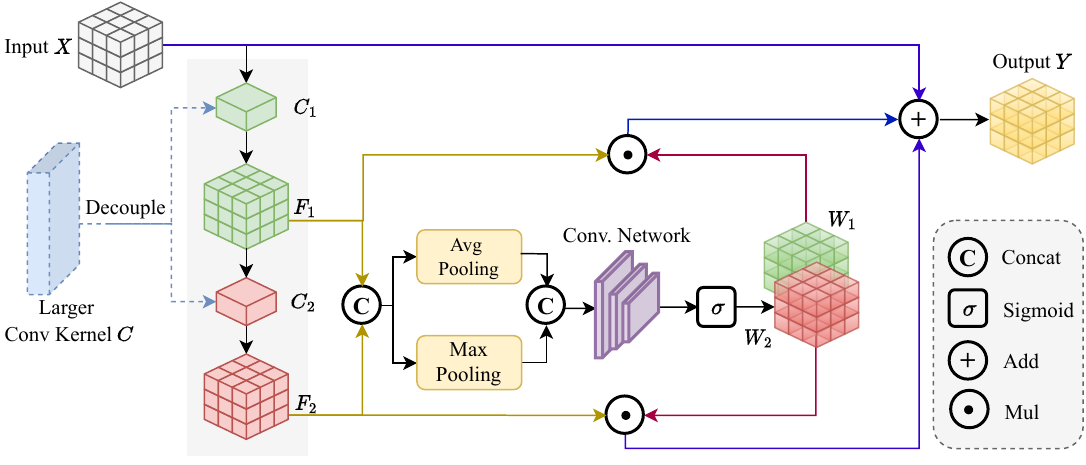}
      \vspace{-1em}
      \caption{DFFM Components: (\textbf{a}) Convolutional decoupling module decomposes large receptive field kernels into smaller ones. (\textbf{b}) Adaptive perception module dynamically adjusts weights of intermediate features across various receptive fields.}
      \vspace{-1.5em}

      \label{fig3.1}
   \end{figure*}
The LargeKernel3D\cite{LargeKernel} introduces a strategy known as local convolution unit weight sharing. In dealing with larger 3D convolution kernels, the convolution units $k \in K_{local}$ located at local positions share identical weights. Through weight sharing across spatial locations, this approach reduces the original large kernel convolution size from 7$\times$7$\times$7 to 3$\times$3$\times$3.\par
Larger kernels cover more extensive processing areas, and accelerating large kernel convolution operations involves multiplexing overlapping regions. LinK\cite{Link} addresses this by introducing a linear kernel. This method maximizes the utilization of global coordinates and feature representations of overlapping blocks, ensuring efficient feature reuse in overlapping regions.

\begin{figure*}[thpb]
      \centering
    \includegraphics[width=0.7\textwidth]{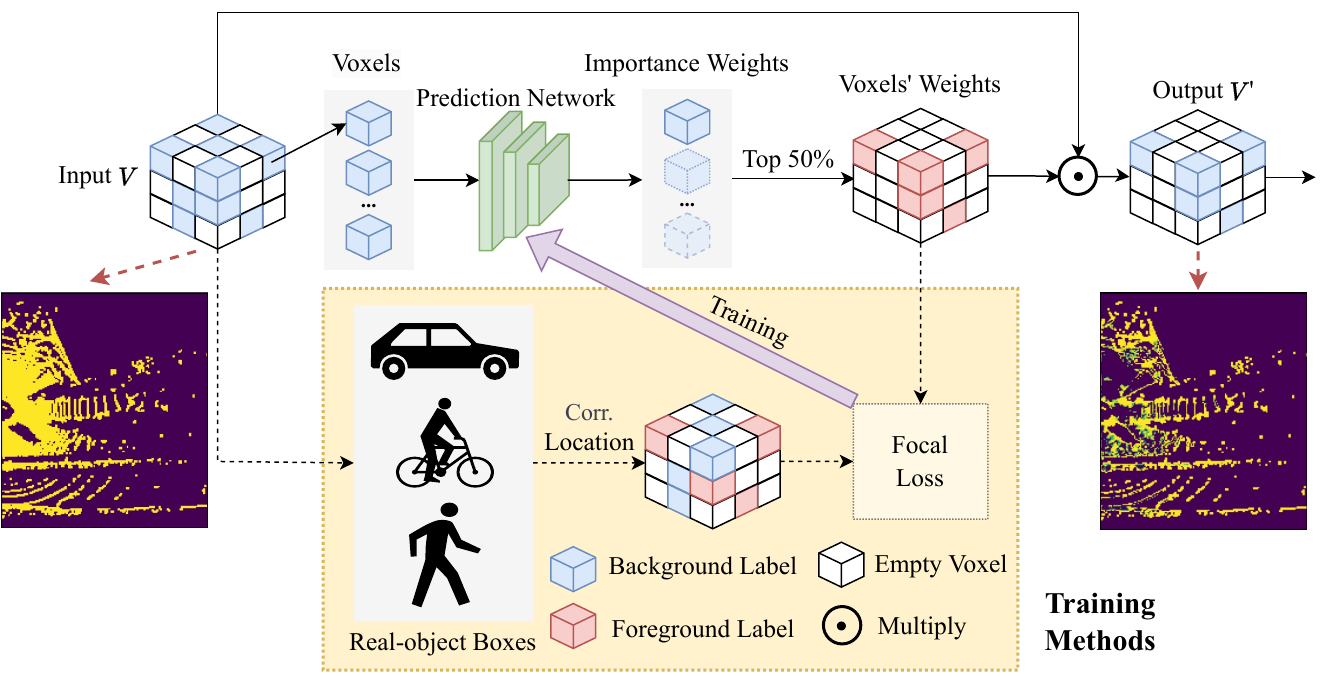}
      \vspace{-1em}
      \caption{The general structure of the FSM. (\textbf{a}) Importance weights are predicted for each voxel using the network. (\textbf{b}) The top 50\% importance weight voxels are retained, discarding the rest. (\textbf{c}) Retained voxel features are multiplied by their weights to create output features. (\textbf{d}) The detection network categorizes samples as positive or negative based on truth box inclusion and adjusts the training accordingly.}
      \vspace{-1.5em}
      \label{fig4.1}
\end{figure*}

\subsection{Important Feature Extraction Strategy}
\label{chap2:c}
CasA\cite{CasA} introduces a cascading attention detector. In contrast to most detectors using a single network for candidate region generation, CasA employs a progressive approach by utilizing a series of sub-networks to extract essential features for prediction boxes. Simultaneously, candidate boxes undergo total refinement by integrating crucial features from different stages, utilizing cascading attention to achieve higher-quality prediction results.\par
A series of studies \cite{PETR, BEVFormer}, exemplified by DETR3D\cite{DETR3D}, employs a decoding detector based on the Transformer architecture. This approach initially defines a set of 3D object queries, each corresponding to the essential features of an output box. These queries engage with the feature map through the attention mechanism to extract feature information corresponding to candidate boxes. The continuous optimization of queries facilitates obtaining their corresponding crucial features.

\section{METHOD}
Firstly, Sec. \ref{sectionA} details the DFFM, specifically designed to accommodate a large receptive field. Subsequently, Sec. \ref{sectionB} introduces the FSM, proficient in quantitative feature filtering. These two modules offer crucial components for enhancing the performance of 3D object detection from distinct perspectives.
Finally, these modules are incorporated into the enhanced network discussed in Sec. \ref{sectionC}.
\subsection{Dynamic Feature Fusion Module}
\label{sectionA}
LargeKernel3D\cite{LargeKernel} effectively reduces the parameter number but falls short in reducing the total atomic operations, which remain tied to the cubic growth of the kernel size. Additionally, its stringent assumption of weight sharing limits the flexibility of the convolution operation.
In contrast, LinK\cite{Link} maps features and coordinates to the range $\left[-1, 1 \right]$, constraining the variation range. The predefined positional deviation further hampers flexible weight learning compared to the conventional convolution.\par
Most crucially, the above method is statically fixed for the extension range of the perceptual field, lacking dynamic adjustment based on the actual required field of view. For instance, a convolutional kernel of size 7$\times$7$\times$7 might cover a vehicle, but it could introduce interfering information when applied to pedestrian detection. This static, fixed sensing field range may not adapt adequately to different scenarios.\par
\textbf{Overall Architecture.} In order to overcome the defects of the above methods, we propose the DFFM, depicted in Fig. \ref{fig3.1}. It consists of the convolutional decoupling and adaptive perception modules. This strategy showcases the ability to leverage intermediate features as a bridge, dynamically expanding the sensing field based on actual requirements while upholding acceptable resource consumption. \par
\textbf{Convolutional Decoupling Module.} The receptive field $RF_i$ of the convolutional kernel of the $i$th layer is calculated as follows:
\begin{eqnarray}
	\label{eq3.7}
	\begin{aligned}
		RF_1 &= k_1,\\
		RF_i &= RF_{i-1}+d_i s_i (k_i - 1).
	\end{aligned}
\end{eqnarray}
Here, $d_i$ represents the dilation rate of the convolution, and $s_i$ is the step size of the convolution-both typically set to 1 in DFFM. Additionally, $k_i$ denotes the size of the convolution kernel.
This formulation enables the decomposition of a large convolution kernel into a collection of convolution kernels $C=\{C_1, C_2, \ldots, C_i \}$-both possessing equivalent receptive fields. Tab. \ref{tab3.4} shows that this decomposition significantly reduces Floating Point Operations (FLOPs) during training.\par
\begin{table}[htbp]
	\centering
	\caption{FLOPs corresponding to different structures.}
	\label{tab3.4}
	\vspace{-0.5em}
	\begin{tabular}{ccc}
		\hline
  
        \hline
		{RF}
		& {sequence ($k_i$, $d_i$)}
		& {FLOPs (M)} \\
		\hline
		\multirow{2}{*}{{5}} & (5, 1) & 91.20 \\
		& (3, 1) $\rightarrow$ (3, 1) & \textbf{79.90} \\
		\hline
  
        \hline
	\end{tabular}
\end{table}
The input features are denoted by $X$, and the intermediate features of the $i$th layer are expressed as:
\begin{eqnarray}
	\label{eq3.8}
		F_1 = C_1(X),\ \cdots,\ F_i = C_i(F_{i-1}).
\end{eqnarray}
Next, we will elaborate on how the adaptive perception module utilizes these intermediate features to construct flexible receptive fields that can be effectively adapted to different scenarios.\par
\textbf{Adaptive Perception Module.} The adjustment of intermediate features at various receptive field sizes is crucial for the dynamic adaptation of the model's receptive field. 
Firstly, we concatenate features at different receptive field sizes:
\begin{eqnarray}
	F = [F_1; F_2; \ldots; F_i].
\end{eqnarray}
Then, channel average pooling $P_{avg}$ and maximum pooling $P_{max}$ are applied to these features, facilitating the effective fusion of the same positional weights in different features:
\begin{eqnarray}
	\label{eq3.10}
	\begin{aligned}
		W_{avg} = P_{avg}(F),\
		W_{max} = P_{max}(F).
	\end{aligned}
\end{eqnarray}
Following this, we employ Eq. \ref{eq3.11} to derive the set of weights $W = [W_1; W_2; \ldots; W_i]$ corresponding to different receptive fields, with $i$ denoting the number of feature maps:
\begin{eqnarray}
	\label{eq3.11}
	W = \sigma (C^{2\rightarrow i}([W_{avg}; W_{max}])).
\end{eqnarray}
In this context, $C^{2\rightarrow i}(\cdot)$ signifies the convolution operation, transitioning from 2 dimensions to $i$ dimensions, while $\sigma(\cdot)$ represents the weight normalization through the Sigmoid activation function. Ultimately, we apply the weights $W_n$ to the intermediate features $F_n$ and subject them to convolution $C_{out}$, yielding the output $Y$.
\begin{eqnarray}
	\label{eq3.12}
	Y = C_{out}(\sum_{n=1}^i(W_n \cdot F_n)) + X.
\end{eqnarray}

\subsection{Feature Selection Module}
\label{sectionB}
We observe that the previous research outlined in Sec.~\ref{chap2:c} primarily emphasizes enhanced feature extraction precision by the design of the detector. However, these methods fall short of effectively minimizing the overall quantity of features. Instead, they rely on the detector's capacity to identify and actively exclude certain invalid features. This strategy not only imposes greater demands on the detector's performance but also introduces redundancy between the detector and the preceding feature extraction network.\par
Consequently, we introduce the Feature Selection Module to diminish irrelevant features, thereby alleviating the computational burden and enabling the detector to concentrate on essential features.\par
\textbf{Feasibility of Feature Filtering.} 
In initial experiments, we assessed the viability of feature filtering through a simple and intuitive method. Our approach involved applying a masking operation to the feature maps. Specifically, we randomly selected a proportions of features from these maps and supplied them to the detector, discarding the remaining features. Encouragingly, the experimental results indicate that the detector's performance is not substantially compromised; instead, there is some degree of improvement. This observation initially validates the redundancy of existing features and underscores the effectiveness of our proposed feature filtering approach.\par
\textbf{Overall Architecture.} 
Many tasks are linked to object detection intricately. Thus, several methods incorporate auxiliary tasks to augment spatial features and offer implicit guidance to achieve high-precision 3D object detection.\par
\begin{figure*}[thpb]
      \centering
    \includegraphics[width=0.78\textwidth]{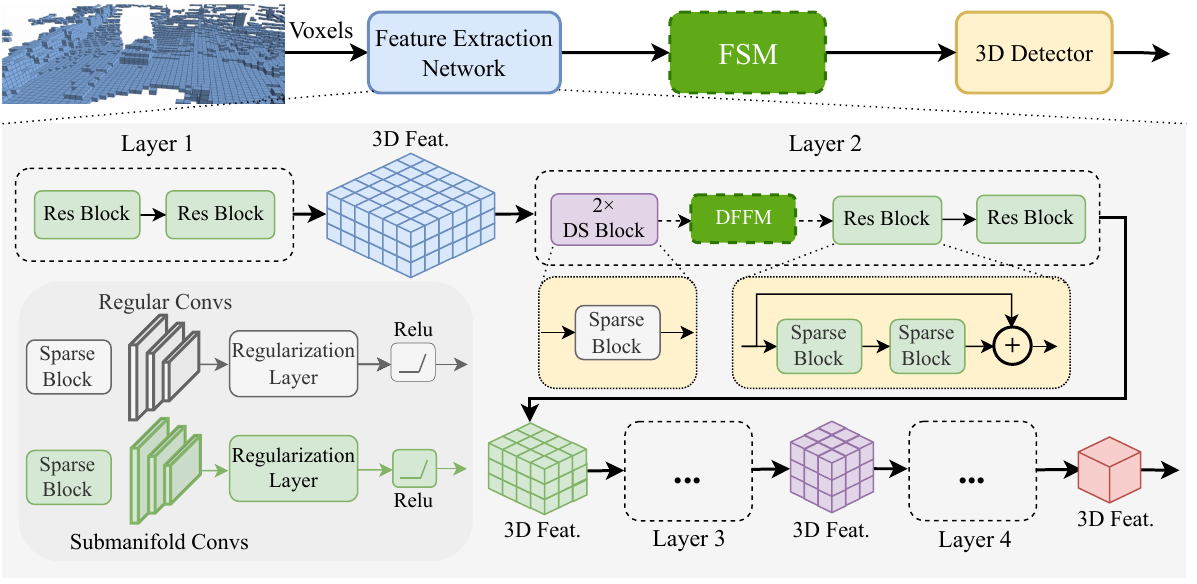}
      \vspace{-1em}
      \caption{Network components: feature extraction network, FSM, and 3D detector. (\textbf{a}) Derived from SECOND\cite{Second}, the feature extraction network has four layers. The first layer includes two residual blocks, and additional downsampling blocks exist the remaining three layers for feature size reduction. (\textbf{b}) Enhancements include DFFM and FSM, added to the position indicated by the dotted line.}
      
      \vspace{-1.5em}

      \label{fig4.2}
\end{figure*}
To assess the significance of features precisely, we introduce prior knowledge that foreground points hold greater importance than background points. To this end,  as illustrated in Fig. \ref{fig4.1}, we devise an additional importance prediction branching task for predicting the importance weights for each voxel grid. \par
Specifically, the voxel features $f_{\mathcal{V}}\in \mathbb{R}^{n_x \times n_y \times n_z \times c}$ are processed by a prediction network $\operatorname{Prediction}^{c\rightarrow 1}(\cdot)$ to obtain the importance weight matrix $W \in [0, 1]^{n_x \times n_y \times n_z}$ of the voxels, as shown in Eq. \ref{eq4.1}.
\begin{eqnarray}
	\label{eq4.1}
	W =\operatorname{Prediction}^{c\rightarrow 1}(f_{\mathcal{V}}).
\end{eqnarray}
Following this, the system will retain the voxel grids $\mathcal{V}_i$ that rank in the top 50\% of the importance weights. The remaining non-empty grids will be set to null, i.e.:
\begin{eqnarray}
	\label{eq4.2}
	\mathcal{V}' = \{\mathcal{V}_i \mid w_i \geq \sigma, \mathcal{V}_i \in \mathcal{V}\},
\end{eqnarray}
where $\mathcal{V}'$ represents the new voxels obtained after the filtering, and $\sigma$ is the threshold for ranking the top 50\% importance weights. Eventually, these retained voxel features $\mathcal{V}_i$ will be multiplied by the corresponding importance weights $w_i$ to form the output features $f_{\mathcal{V}'}$.
\begin{eqnarray}
	\label{eq4.3}
	f_{\mathcal{V}'} =w  \odot f_{\mathcal{V}}.
\end{eqnarray}\par
During training, the prediction network treats the voxels within the truth box as positive samples while the remaining are negative. Focal Loss\cite{Focal-loss} is utilized as the loss function, and parameters are adjusted through backpropagation.\par
This strategy incorporates model compression by diminishing unimportant features and accentuating critical ones. The subsequent detectors can then prioritize learning the most essential features. Simultaneously, the reduction of empty anchor frames mitigates mutual interference between candidate frames. This approach significantly diminishes the computational load and enhances the network's convergence speed and processing efficiency during training.
\subsection{Overall Network Structure}
\label{sectionC}
The SECOND network \cite{Second} adopts a 3D sparse convolutional network structure, which has become the standard for voxel feature extraction networks. Specifically, the network comprises four layers.
In the first layer, raw voxels undergo processing by a residual block consisting of two submanifold sparse convolutions. Each subsequent layer initially employs a 2$\times$ regular sparse convolution downsampling block to halve the feature size before engaging in feature extraction using the residual block. The same structure is used for many subsequent works \cite{PDV,PV-rcnn,PV-rcnn++,Voxel-rcnn,VoxelNext}.\par
We enhance the baseline network by integrating the DFFM into the feature extraction network, as illustrated in Fig. \ref{fig4.2}. Subsequently, the FSM is added before the detector. To validate the effect of our strategy, we implement the improved network on SECOND\cite{Second} and VoxelNext\cite{VoxelNext}.
\section{EXPERIMENTS}
\subsection{Dataset}
KITTI \cite{KITTI} serves as a widely recognized benchmark for autonomous driving datasets. It encompasses three categories: car, cyclist, and pedestrian, each further classified into three difficulty levels—easy, medium, and hard—based on object size, occlusion level, and truncation level. The dataset consists of 7481 training samples and 7518 test samples. For validation, the training samples are commonly split into a training set of 3712 and a validation set of 3769. \par
\begin{table*}[h]
\caption{3D object detection results on the KITTI validation set. The results are evaluated with AP$|_{R_{40}}$.}

\label{table2}
\begin{center}
\renewcommand{\arraystretch}{1.2}
\vspace{-1em}
\begin{tabular}{ccccccccccc}
\hline

\hline
\multirow{2}{*}{Method}
& \multicolumn{3}{c}{Car}
& \multicolumn{3}{c}{Cyclist}    
& \multicolumn{3}{c}{Pedestrian} 
&\multirow{2}{*}{3D mAP}
\\
\cline{2-10}
&Easy&Mod.&Hard&Easy&Mod.&Hard&Easy&Mod.&Hard\\
\hline
SECOND\textsuperscript{\dag}\cite{Second}  & 90.11 & 81.08 & 78.11
& 85.53 & 68.58 & 64.45 & 57.67 & 51.92 & 47.02 & 69.38\\
DFFM+SECOND & \textbf{90.42}\textsuperscript{\ddag} & \textbf{81.37} & \textbf{78.52} & \textbf{86.11} & \textbf{68.86} & \textbf{64.55} & \textbf{58.70} & \textbf{53.15} & \textbf{49.12} & \textbf{70.09}\\ 
\rowcolor{green!10}Improvement& +\textit{0.31} & +\textit{0.29} & +\textit{0.41} & +\textit{0.58} & +\textit{0.28} & +\textit{0.10} & +\textit{1.03} & +\textit{1.23} & +\textit{2.10} & +\textit{0.71}\\
FSM+SECOND & \textbf{90.25} & \textbf{81.64} & \textbf{78.75} & \textbf{86.29} & 68.58 & 64.15 & \textbf{57.87} & \textbf{53.32} & \textbf{49.16} & \textbf{70.17}\\ 
\rowcolor{red!10}Improvement& +\textit{0.14} & +\textit{0.56} & +\textit{0.64} & +\textit{0.76} & +\textit{0.00} & {-\textit{0.30}} & +\textit{0.20} & +\textit{1.40} & +\textit{2.14} & +\textit{0.79}\\
\hline
VoxelNext\textsuperscript{\dag}\cite{VoxelNext} & 86.89 & 78.00 & 75.50
		& 87.28 & 68.76 & 65.71 & 61.46 & 55.51 & 50.76 & 69.99\\
DFFM+VoxelNext & \textbf{88.92} & \textbf{80.48} & \textbf{78.09} & \textbf{90.56} & \textbf{70.03} & \textbf{65.89} & \textbf{63.79} & \textbf{58.23} & \textbf{53.33} & \textbf{72.15} \\
\rowcolor{green!10}Improvement& +\textit{2.03} & +\textit{2.48} & +\textit{2.59} & +\textit{3.28} & +\textit{1.27} & +\textit{0.18} & +\textit{2.33} & +\textit{2.72} & +\textit{2.57} & +\textit{2.12}\\
FSM+VoxelNext & 86.24 & 74.59 & 72.34 & \textbf{88.25} & \textbf{70.24} & \textbf{66.12} & \textbf{69.04} & \textbf{63.05} & \textbf{57.61} & \textbf{71.94} \\
\rowcolor{red!10}Improvement& -\textit{0.65} & -\textit{3.41} & -\textit{3.16} & +\textit{0.97} & +\textit{1.48} & +\textit{0.41} & +\textit{7.58} & +\textit{7.54} & +\textit{6.85} & +\textit{1.95}\\
\hline

\hline
\end{tabular}
\end{center}
\noindent{\footnotesize{\textsuperscript{\dag} The detection results of the benchmark are reproduced by its  official released code.}}

\noindent{\footnotesize{\textsuperscript{\ddag} The improved results are in bold.}}
\vspace{-2em}
\end{table*}
The official evaluation criterion of the KITTI dataset is AP$|_{R_{40}}$.
The KITTI dataset uses AP$|_{R_{40}}$  as its official evaluation criterion. It first splits the horizontal coordinate into forty equal parts corresponding to the recall points $r_i \in R = \{1/40, 2/40, \dots, 1\}$. Next, the Precision-Recall curve is smoothed by setting the recall value $\operatorname{P}'(r_i)$ to the largest recall value $\operatorname{P}(r')$ on the right. The value of AP$|_{R_{40}}$ is calculated by the area under the smoothed curve, as:
\begin{eqnarray}
	\label{eq3.13}
	\begin{aligned}
	\operatorname{AP}|_{R_{40}}&=\frac{1}{|40|} \sum_{r_i \in R} \operatorname{P}'(r_i), \\
		\operatorname{P}'(r_i)&=\max \operatorname{P}(r') \quad \text{subject to } r^{\prime} \geq r_i.
	\end{aligned}
\end{eqnarray}
\subsection{Setup Details}

\textbf{Data augmentation.} 
Data augmentation is an effective strategy for diversifying training data and thus improving  model's generalization ability. Following \cite{Second}, we extract some ground truth boxes and place them randomly. Then, point clouds undergo random transformations, including flipping along the $x$-axis, rotation along the $z$-axis in the range of [-$\pi$/4, $\pi$/4], and scaling within the range [0.95, 1.05].\par

\begin{table}[h]
\setlength{\abovecaptionskip}{0cm}
\setlength{\belowcaptionskip}{10pt}
\caption{Effects of different method on the KITTI validation set.}
\vspace{-0.5em}
\label{tableadd1}
\begin{center}
\renewcommand{\arraystretch}{1.2}
\begin{tabular}{cccc}
\hline

\hline
Method &Baseline\cite{Second} & LargeKernel3D\cite{LargeKernel} & DFFM \\
\hline
3D mAP(\%) &69.38 & 69.84& \textbf{70.09} \\
\hline

\hline
\end{tabular}
\end{center}
\vspace{-2.5em}
\end{table}

\textbf{Input Parameters.}
The KITTI dataset provides annotated information only for objects in the field of the front camera's view. Therefore, we constrain the point clouds along the $(x, y, z)$ axes to [0, 70.4m], [-40m, 40m] and [-3m, 1m], respectively. We also adjust the voxel size to (0.05m, 0.05m, 0.1m). Consequently, the size of the entire 3D space after voxelization is 1600$\times$1408$\times$40. The maximum number of non-empty voxels is set to 16000 in the training set and 40000 in the test set.\par
\textbf{Training.}
We use OpenPCDet\cite{OpenPCDet} as the code library for our development. All models are trained on two 3090Ti graphics cards, using a Batch Size setting of 4, for 80 training epochs. Following the optimization strategy of the baseline detector \cite{Second}, we train the models using the Adam optimizer with an initial learning rate of 0.003, Momentum of 0.9, and a weight decay parameter of 0.01.
\subsection{Ablation Studies}
We first evaluate our strategies using SECOND\cite{Second} and VoxelNext\cite{VoxelNext} as benchmark models, representing voxel and sparse detection networks, respectively.\par
\textbf{DFFM.} As indicated in Tab. \ref{table2}, the incorporation of DFFM enhances the overall 3D mAP performance of the SECOND by 0.71\% and the VoxelNext network by an astonishing 2.12\%. Notably, DFFM improves the model's detection performance for pedestrian by (1.03\%, 1.23\%, 2.10\%) and (2.33\%, 2.72\%, 2.57\%), respectively, on AP$|_{R_{40}}$. Tab. \ref{tableadd1} also shows DFFM outperforms the similar method LargeKernel3D \cite{LargeKernel} in overall performance.\par

\begin{figure}[thpb]
      \centering
      \vspace{-1em}
      \includegraphics[width=0.38\textwidth]{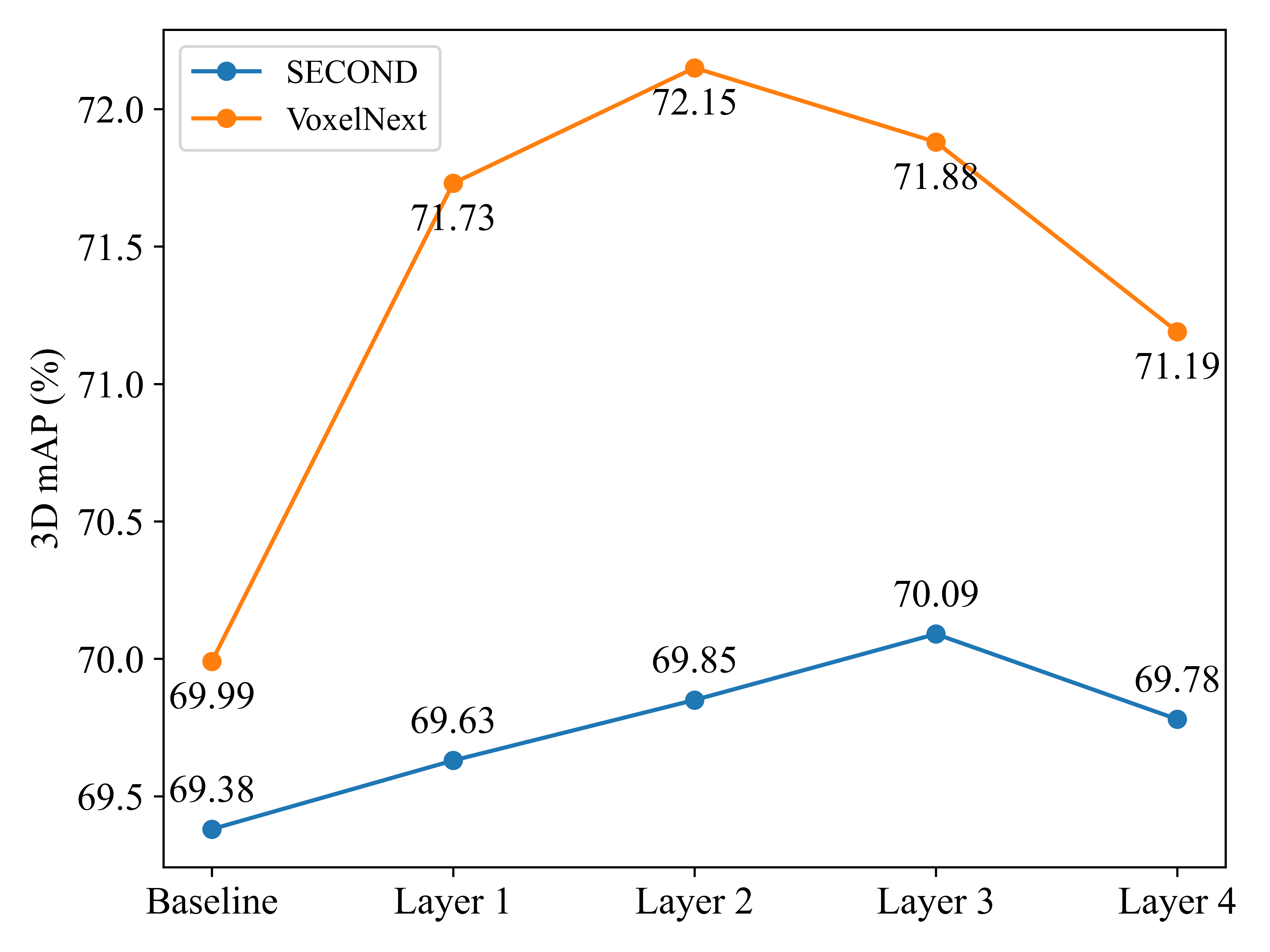}
      \vspace{-1em}
      \caption{Effect of DFFM in different stages.}
      \vspace{-1.5em}
      \label{fig5-1}
\end{figure}

We further conduct a comparative experiment on the effect of integrating DFFM at different stages. As shown in Fig. \ref{fig5-1}, no matter which stage the DFFM is placed in, it exhibits a certain degree of performance improvement compared to the original network. Notably, the most substantial enhancements are observed when DFFM is set into the middle stage. Moreover, the network's performance exhibits a trend of increase and decrease with the increase of the placement stage.\par

\textbf{FSM.} The data in Tab. \ref{table2} indicate that after applying FSM, the 3D mAP of SECOND and VoxelNext improves by 0.79\% and 1.95\%, respectively. The SECOND detector exhibits enhanced detection across various objects, and the feature-filtered detector demonstrates improved detection of small objects, particularly in VoxelNext, the detection performance of pedestrian with different difficulties significantly improves by (7.58\%, 7.54\%, and 6.85\%). \par
The VoxelNext detector experiences a reduction in vehicle detection performance after applying the FSM. We believe this is not due to the FSM mistakenly discarding essential features, but rather because of challenges posed by the vehicle's large size and the limited receptive field of the feature extraction network. This limitation impedes the learning of distinct features between important voxels inside the vehicle and the background voxels. Consequently, the FSM leads to the loss of front-back hierarchical information, making it difficult for the detector to accurately delineate the boundaries of the real box. The following overall experiment also verified this.\par

\begin{table}[h]
\setlength{\abovecaptionskip}{0cm}
\setlength{\belowcaptionskip}{10pt}
\vspace{-1em}
\caption{Average time for single-frame inference across networks on the KITTI validation set.}
\vspace{-1em}
\label{table3}
\begin{center}
\renewcommand{\arraystretch}{1.2}
\begin{tabular}{cccc}
\hline

\hline
Method & Average time (ms) & FPS &Improvement\\
\hline
SECOND\cite{Second} & 69.59 &14.37& - \\
		SECOND+FSM & \textbf{63.10} &\textbf{15.85} & \textbf{9.33\%} \\
		\hline VoxelNext\cite{VoxelNext} & 51.73 & 19.33&- \\
		VoxelNext+FSM & \textbf{41.60} &  \textbf{24.04}&\textbf{19.56\%} \\
\hline

\hline
\end{tabular}
\end{center}
\vspace{-1.5em}
\end{table}

\begin{table}[h]
\caption{3D object detection results on the KITTI validation set. The results are evaluated with AP$|_{R_{11}}$.}

\label{table-1}
\begin{center}
\renewcommand{\arraystretch}{1.2}
\vspace{-2em}
\begin{tabular}{cccc}
\hline

\hline
Method
& Car
& Cyclist
& Pedestrian
\\
\hline
PointPillar\cite{Pointpillars}& 77.28 & 62.68 & 52.29\\
SECOND-IoU & 79.09 & 71.31 & 55.74\\
PointRCNN\cite{Pointrcnn} & 78.70 & 72.11 & 54.41\\
Part-$A^2$\cite{PartA2} &79.40 & 69.90 & \textbf{60.05}\\
PV-RCNN\cite{PV-rcnn} & 83.61 & 70.47 & 57.90\\
Focals Conv\cite{focal} & 85.66 & - & - \\
\rowcolor{green!10}\textbf{Ours} & \textbf{86.62} & \textbf{79.23} & 59.75 \\
\hline

\hline
\end{tabular}
\end{center}
\end{table}

\begin{figure}[thpb]
      \centering
      \vspace{-1em}
      \includegraphics[width=0.38\textwidth]{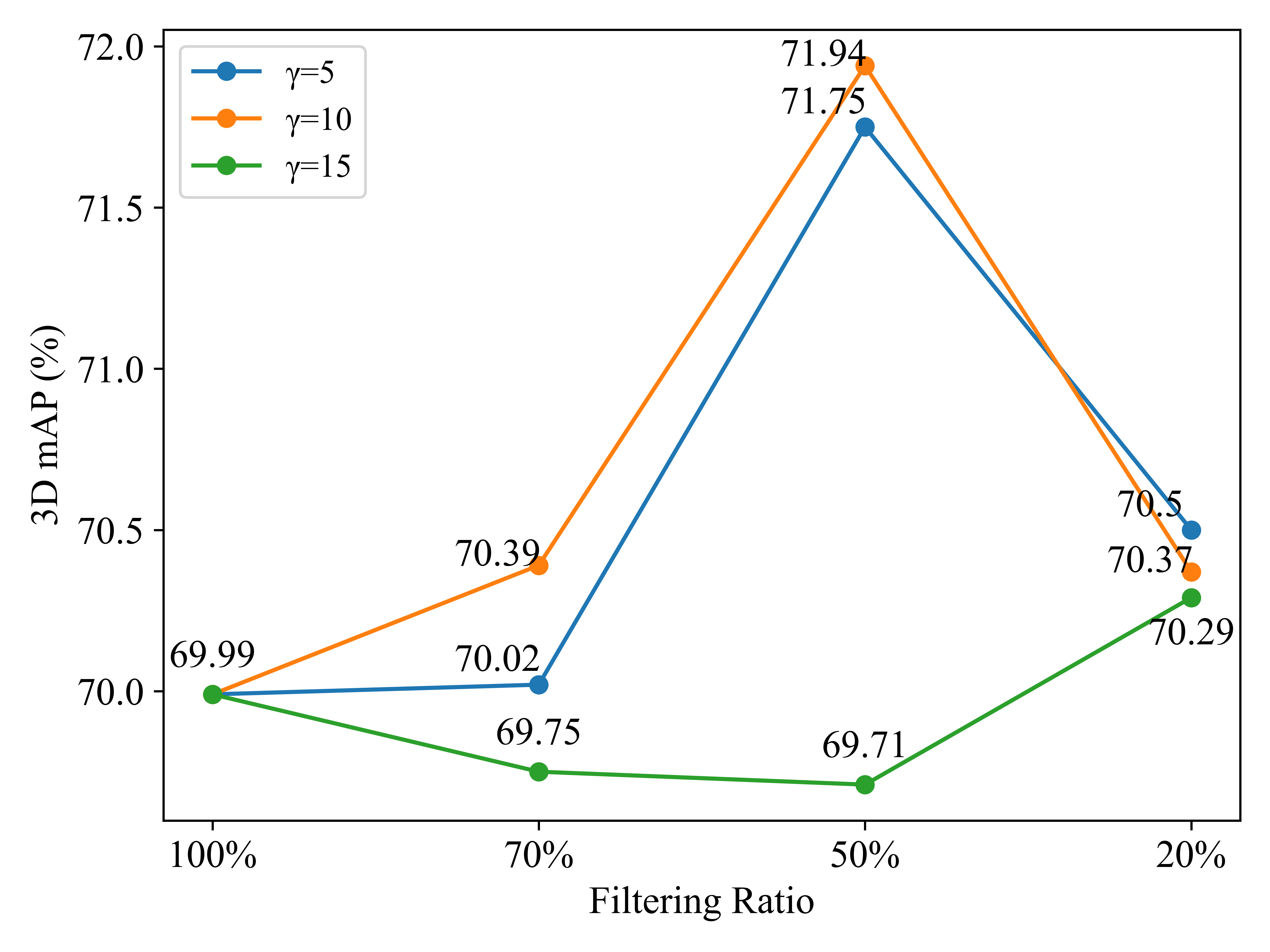}
      \vspace{-1em}
      \caption{Effect of FSM with different settings.}
      \vspace{-1.5em}
      \label{fig5-2}
\end{figure}

As shown in Tab. \ref{table3}, the FSM reduces the inference time of SECOND and VoxelNext by 9.33\% and 19.56\%, effectively improving the models' real-time running speed.\par
We use VoxelNext as the benchmark and adjust the importance filtering ratio along with the weight $\gamma$ of the importance prediction loss. In Fig. \ref{fig5-2}, when $\gamma=15$, the loss weight is excessively large, hindering the training of the detection and yielding poor results. However, with $\gamma=5$ or $\gamma=10$, the ratio is reduced within a specific range, leading to decreased interference between candidate boxes and improved detector performance. Although the lower ratio excludes many foreground voxels, resulting in a slight performance degradation, it still surpasses the original network, providing further evidence of feature redundancy.\par
\textbf{Both.} Our network outperforms state-of-the-art networks in car and cyclist according to the results in Tab. \ref{table-1}. Although slightly trailing behind in pedestrian, our performance remains competitive.\par
The combination of DFFM and FSM enhances the network performance, as shown in Tab. \ref{table4}. This not only validates the effectiveness of DFFM but also underscores its complementary role with FSM. The dynamic receptive field enables flexible extraction of required information from various classes of voxels. Subsequent feature filtering directly eliminates unimportant voxels without compromising front and back hierarchical information, leading to improved model performance.\par
\begin{table}[h]
\setlength{\abovecaptionskip}{0cm}
\setlength{\belowcaptionskip}{10pt}
\vspace{-1em}
\caption{Effects of proposed components on the KITTI validation set.}
\vspace{-1em}
\label{table4}
\begin{center}
\renewcommand{\arraystretch}{1.2}
\begin{tabular}{ccccc}
\hline

\hline
Baseline & FSM & DFFM &  Two-stage &3D mAP(\%)\\
\hline
\checkmark & & & &69.99 \\
		\checkmark & \checkmark& &  &71.94\\
	 	\checkmark & \checkmark& \checkmark & &\textbf{72.90}\cellcolor{green!10}(+\textit{2.91})\\
    \checkmark & \checkmark& \checkmark & \checkmark &\textbf{75.20}\cellcolor{green!10}(+\textit{5.21}) \\
\hline

\hline
\end{tabular}
\end{center}
\vspace{-2em}
\end{table}

\subsection{Visualizations} 
\begin{figure}[thpb]
      \centering
      \vspace{-1em}
      \includegraphics[width=0.49\textwidth]{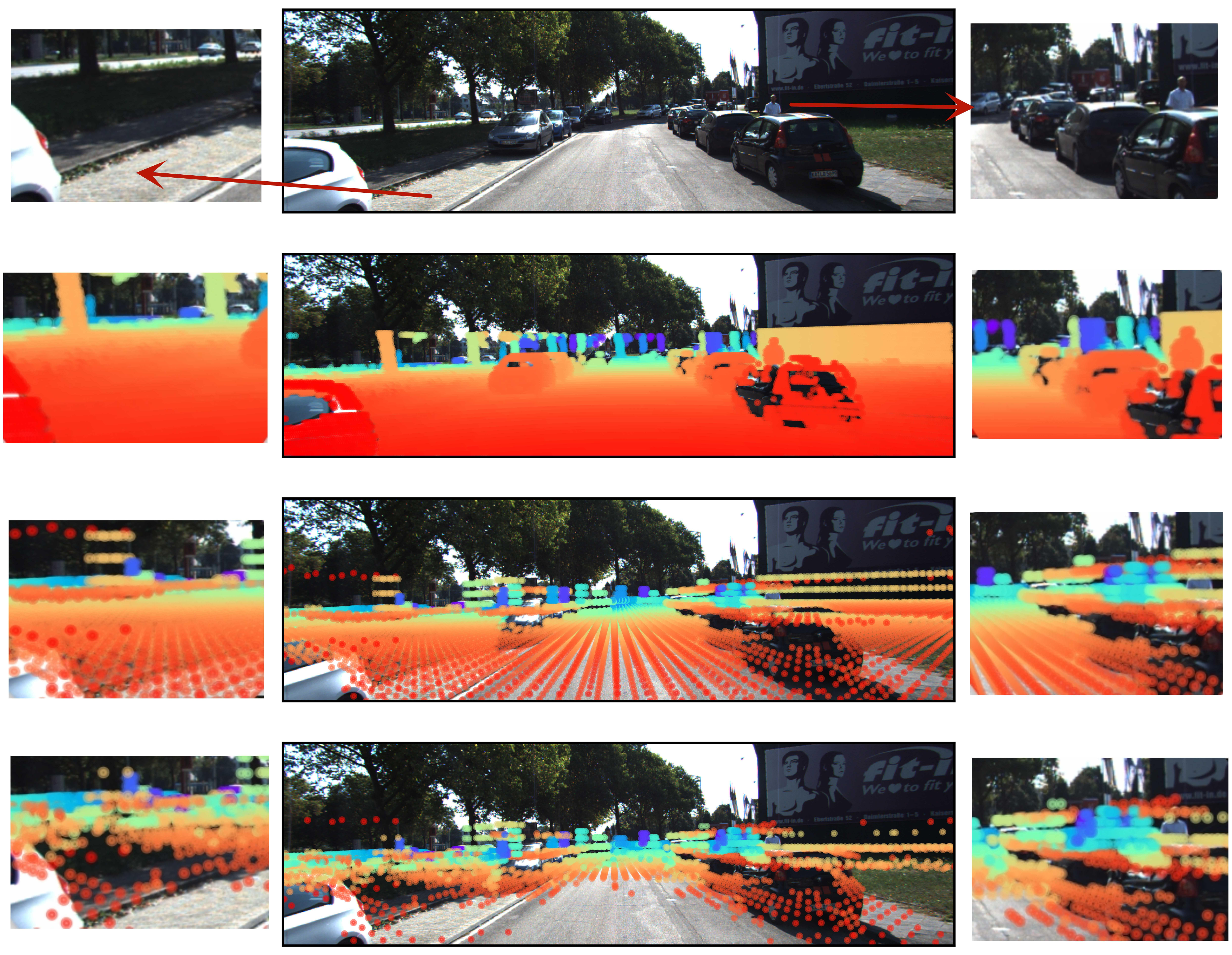}
      \vspace{-2em}
      \caption{Visual representations of various features: sequential images of the original picture, original point clouds mapped to the picture, and voxels before and after FSM processing.}
      \vspace{-1em}
      \label{fig5-3}
\end{figure}
\textbf{Feature Visualization.} We employ geometric relations to map voxels to the image's coordinate system, allowing us to observe changes in voxel features after passing through the FSM. In Fig. \ref{fig5-3}, visualized images progress sequentially from top to bottom, showcasing the original picture, original point clouds mapped to the picture, and voxels mapped to the image before and after going through the FSM. Different colors of points on the picture indicate varying distances.\par
Observations indicate that critical voxels of real objects remain unchanged in the locally enlarged right-column images. In contrast, the left-column images effectively remove unimportant voxels in background elements such as highways, grass, and signage.\par
\begin{figure}[thpb]
      \centering
      \includegraphics[width=0.42\textwidth]{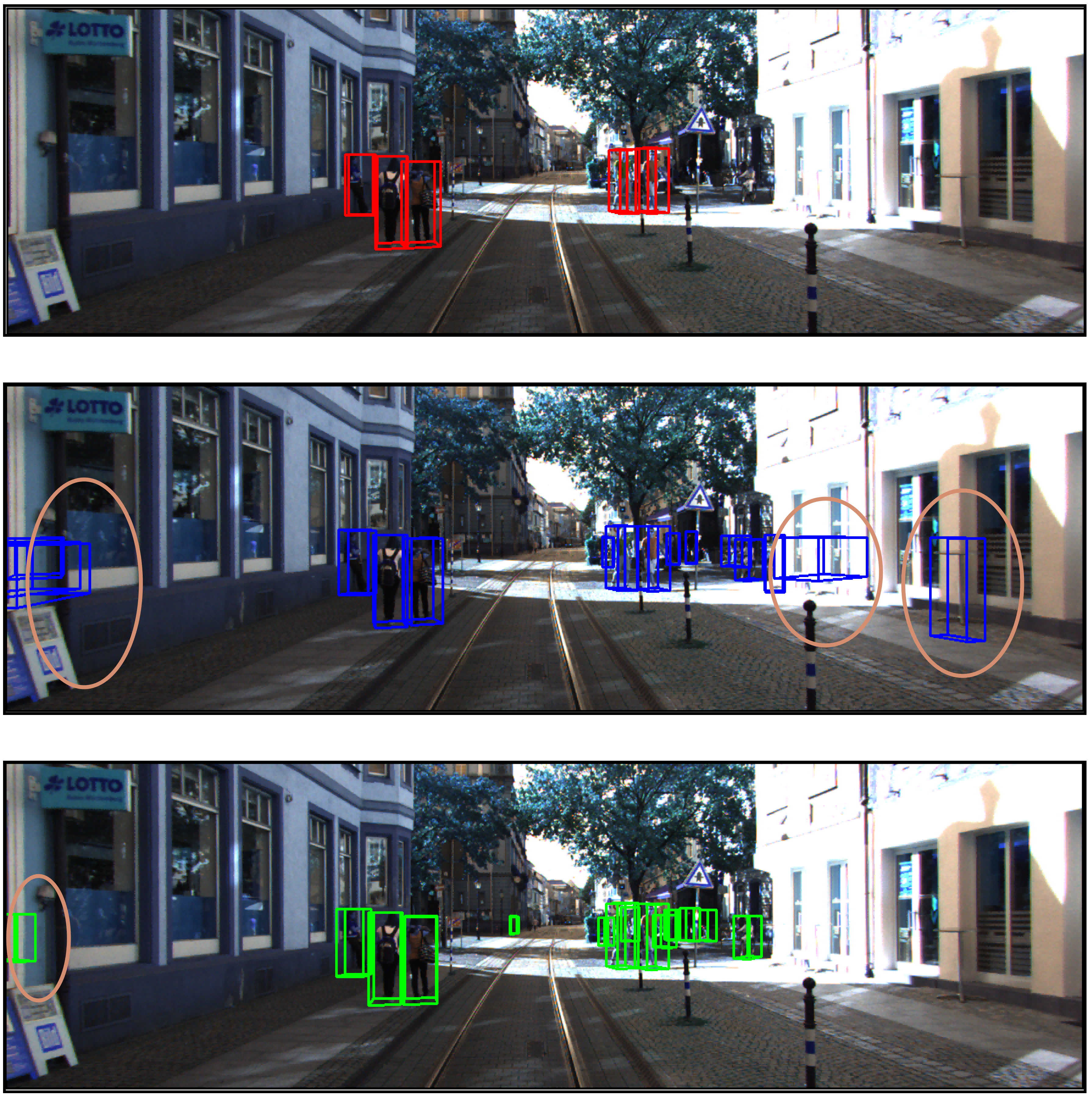}
      \vspace{-1em}
      \caption{Comparing model outputs: true labels (red), original model results (blue), and improved model outcomes (green).}
      \vspace{-1em}
      \label{fig5-4}
\end{figure}

\textbf{Outputs.}
Considering the substantial enhancement in small object detection, we focused on specific scenarios such as people and bicycles, highlighting this progress through a comparative analysis of detection outcomes. In Fig. \ref{fig5-4}, red boxes indicate true labels, blue boxes represent results from the original model, and outcomes from the improved model are depicted with green boxes. Encouragingly, the improved model not only adeptly identifies previously undetected distant pedestrian but also markedly reduces false detections compared to the original model. This robustly substantiates the pivotal role of our strategy in advancing small object detection.
\section{CONCLUSIONS}
In addressing the challenges associated with 3D target detection, this article introduces two pivotal modules: the Dynamic Feature Fusion Module (DFFM) and the Feature Selection Module (FSM). These modules are dedicated to achieving large receptive field and extracting important features, respectively, presenting a novel approach to large receptive field and important feature extraction strategies.
To overcome the challenge of expanding the receptive field of a 3D convolutional kernel, our proposed DFFM introduces an adaptive approach. This module effectively balances the expansion of the 3D convolutional kernel's receptive field with acceptable computational loads, reducing operations while enabling dynamic adjustments to different object requirements.
Simultaneously, the FSM identifies and eliminates redundant information in 3D features. The FSM segregates output box fitting from feature extraction through quantitative evaluation and discarding of unimportant features. This leads to model compression, decreasing computational burden, and mitigating interference among candidate boxes.
Our extensive experiments demonstrate the effectiveness of both DFFM and FSM, particularly in improving small target detection and accelerating network performance. Importantly, these modules exhibit effective complementarity, showcasing their combined impact on advancing the state-of-the-art in 3D object detection.
\addtolength{\textheight}{-0.5cm}   

\bibliographystyle{IEEEtran}
\bibliography{cite}

\begin{thebibliography}{10}
\providecommand{\url}[1]{#1}
\csname url@samestyle\endcsname
\providecommand{\newblock}{\relax}
\providecommand{\bibinfo}[2]{#2}
\providecommand{\BIBentrySTDinterwordspacing}{\spaceskip=0pt\relax}
\providecommand{\BIBentryALTinterwordstretchfactor}{4}
\providecommand{\BIBentryALTinterwordspacing}{\spaceskip=\fontdimen2\font plus
\BIBentryALTinterwordstretchfactor\fontdimen3\font minus
  \fontdimen4\font\relax}
\providecommand{\BIBforeignlanguage}[2]{{%
\expandafter\ifx\csname l@#1\endcsname\relax
\typeout{** WARNING: IEEEtran.bst: No hyphenation pattern has been}%
\typeout{** loaded for the language `#1'. Using the pattern for}%
\typeout{** the default language instead.}%
\else
\language=\csname l@#1\endcsname
\fi
#2}}
\providecommand{\BIBdecl}{\relax}
\BIBdecl

\bibitem{SUY31}
X.~Ding, X.~Zhang, J.~Han, and G.~Ding, ``{Scaling up your kernels to 31x31:
  Revisiting large kernel design in CNNs},'' in \emph{CVPR}, 2022, pp.
  11\,963--11\,975.

\bibitem{ConvNext}
Z.~Liu, H.~Mao, C.-Y. Wu, C.~Feichtenhofer, T.~Darrell, and S.~Xie, ``{A
  ConvNet for the 2020s},'' in \emph{CVPR}, 2022, pp. 11\,976--11\,986.

\bibitem{VoxelNext}
Y.~Chen, J.~Liu, X.~Zhang, X.~Qi, and J.~Jia, ``{VoxelNext: Fully sparse
  VoxelNet for 3D object detection and tracking},'' in \emph{CVPR}, 2023, pp.
  21\,674--21\,683.

\bibitem{Second}
Y.~Yan, Y.~Mao, and B.~Li, ``{SECOND: Sparsely embedded convolutional
  detection},'' \emph{Sensors}, vol.~18, no.~10, p. 3337, 2018.

\bibitem{Pointnet++}
C.~R. Qi, L.~Yi, H.~Su, and L.~J. Guibas, ``{PointNet$++$: Deep hierarchical
  feature learning on point sets in a metric space},'' \emph{ANIPS}, vol.~30,
  2017.

\bibitem{ngiam}
J.~Ngiam, B.~Caine, W.~Han, B.~Yang, Y.~Chai, P.~Sun, Y.~Zhou, X.~Yi,
  O.~Alsharif, P.~Nguyen \emph{et~al.}, ``{StarNet: Targeted computation for
  object detection in point clouds},'' \emph{arXiv preprint arXiv:1908.11069},
  2019.

\bibitem{3DSSD}
Z.~Yang, Y.~Sun, S.~Liu, and J.~Jia, ``{3DSSD: Point-based 3D single stage
  object detector},'' \emph{CVPR}, 2020.

\bibitem{Pointformer}
X.~Pan, Z.~Xia, S.~Song, L.~E. Li, and G.~Huang, ``{3D object detection with
  pointformer},'' in \emph{CVPR}, 2021, pp. 7463--7472.

\bibitem{SASSD}
C.~He, H.~Zeng, J.~Huang, X.-S. Hua, and L.~Zhang, ``{Structure aware
  single-stage 3D object detection from point cloud},'' in \emph{CVPR}, 2020,
  pp. 11\,873--11\,882.

\bibitem{CenterPoint}
T.~Yin, X.~Zhou, and P.~Krahenbuhl, ``{Center-based 3D object detection and
  tracking},'' in \emph{CVPR}, 2021, pp. 11\,784--11\,793.

\bibitem{PartA2}
S.~Shi, Z.~Wang, J.~Shi, X.~Wang, and H.~Li, ``{From points to parts: 3D object
  detection from point cloud with part-aware and part-aggregation network},''
  \emph{TPAMI}, vol.~43, no.~8, pp. 2647--2664, 2020.

\bibitem{PV-rcnn}
S.~Shi, C.~Guo, L.~Jiang, Z.~Wang, J.~Shi, X.~Wang, and H.~Li, ``{PV-RCNN:
  Point-voxel feature set abstraction for 3D object detection},'' in
  \emph{CVPR}, 2020, pp. 10\,529--10\,538.

\bibitem{MMFusion}
L.~Cui, X.~Li, M.~Meng, and X.~Mo, ``{MMFusion: A generalized multi-modal
  fusion detection framework},'' in \emph{IEEE International Conference on
  Development and Learning}, 2023, pp. 415--422.

\bibitem{VeloFCN}
B.~Li, T.~Zhang, and T.~Xia, ``{Vehicle detection from 3D LiDAR using fully
  convolutional network},'' \emph{arXiv preprint arXiv:1608.07916}, 2016.

\bibitem{LMNet}
K.~Minemura, H.~Liau, A.~Monrroy, and S.~Kato, ``{LMNet: Real-time multiclass
  object detection on CPU using 3D LiDAR},'' in \emph{Asia-Pacific Conference
  on Intelligent Robot Systems}.\hskip 1em plus 0.5em minus 0.4em\relax IEEE,
  2018, pp. 28--34.

\bibitem{GraphNet}
M.~Feng, S.~Z. Gilani, Y.~Wang, L.~Zhang, and A.~Mian, ``{Relation graph
  network for 3D object detection in point clouds},'' \emph{IEEE Transactions
  on Image Processing}, vol.~30, pp. 92--107, 2020.

\bibitem{Resnet}
K.~He, X.~Zhang, S.~Ren, and J.~Sun, ``{Deep residual learning for image
  recognition},'' in \emph{CVPR}, 2016, pp. 770--778.

\bibitem{swintransformer}
Z.~Liu, Y.~Lin, Y.~Cao, H.~Hu, Y.~Wei, Z.~Zhang, S.~Lin, and B.~Guo, ``{Swin
  Transformer: Hierarchical vision transformer using shifted windows},'' in
  \emph{ICCV}, 2021, pp. 10\,012--10\,022.

\bibitem{Pointrcnn}
S.~Shi, X.~Wang, and H.~Li, ``{PointRCNN: 3D object proposal generation and
  detection from point cloud},'' in \emph{CVPR}, 2019, pp. 770--779.

\bibitem{Pointpillars}
A.~H. Lang, S.~Vora, H.~Caesar, L.~Zhou, J.~Yang, and O.~Beijbom,
  ``{PointPillars: Fast encoders for object detection from point clouds},'' in
  \emph{CVPR}, 2019, pp. 12\,697--12\,705.

\bibitem{PT}
H.~Zhao, L.~Jiang, J.~Jia, P.~H. Torr, and V.~Koltun, ``{Point Transformer},''
  in \emph{ICCV}, 2021, pp. 16\,259--16\,268.

\bibitem{PCT}
M.-H. Guo, J.-X. Cai, Z.-N. Liu, T.-J. Mu, R.~R. Martin, and S.-M. Hu, ``{PCT:
  Point cloud transformer},'' \emph{Computational Visual Media}, vol.~7, pp.
  187--199, 2021.

\bibitem{Voxel-transformer}
J.~Mao, Y.~Xue, M.~Niu, H.~Bai, J.~Feng, X.~Liang, H.~Xu, and C.~Xu, ``{Voxel
  transformer for 3D object detection},'' in \emph{ICCV}, 2021, pp. 3164--3173.

\bibitem{Swformer}
P.~Sun, M.~Tan, W.~Wang, C.~Liu, F.~Xia, Z.~Leng, and D.~Anguelov, ``{SWFormer:
  Sparse window transformer for 3D object detection in point clouds},'' in
  \emph{ECCV}.\hskip 1em plus 0.5em minus 0.4em\relax Springer, 2022, pp.
  426--442.

\bibitem{LargeKernel}
Y.~Chen, J.~Liu, X.~Zhang, X.~Qi, and J.~Jia, ``{LargeKernel3D: Scaling up
  kernels in 3D sparse CNNs},'' in \emph{CVPR}, 2023, pp. 13\,488--13\,498.

\bibitem{Link}
T.~Lu, X.~Ding, H.~Liu, G.~Wu, and L.~Wang, ``{LinK: Linear kernel for
  LiDAR-based 3D perception},'' in \emph{CVPR}, 2023, pp. 1105--1115.

\bibitem{CasA}
H.~Wu, J.~Deng, C.~Wen, X.~Li, C.~Wang, and J.~Li, ``{CasA: A cascade attention
  network for 3D object detection from LiDAR point clouds},'' \emph{IEEE
  Transactions on Geoscience and Remote Sensing}, vol.~60, pp. 1--11, 2022.

\bibitem{PETR}
Y.~Liu, T.~Wang, X.~Zhang, and J.~Sun, ``{PETR: Position embedding
  transformation for multi-view 3D object detection},'' in \emph{ECCV}.\hskip
  1em plus 0.5em minus 0.4em\relax Springer, 2022, pp. 531--548.

\bibitem{BEVFormer}
Z.~Li, W.~Wang, H.~Li, E.~Xie, C.~Sima, T.~Lu, Y.~Qiao, and J.~Dai,
  ``{BEVFormer: Learning bird’s-eye-view representation from multi-camera
  images via spatiotemporal transformers},'' in \emph{ECCV}.\hskip 1em plus
  0.5em minus 0.4em\relax Springer, 2022, pp. 1--18.

\bibitem{DETR3D}
Y.~Wang, V.~C. Guizilini, T.~Zhang, Y.~Wang, H.~Zhao, and J.~Solomon,
  ``{DETR3D: 3D object detection from multi-view images via 3D-to-2D
  queries},'' in \emph{CRL}.\hskip 1em plus 0.5em minus 0.4em\relax PMLR, 2022,
  pp. 180--191.

\bibitem{Focal-loss}
T.-Y. Lin, P.~Goyal, R.~Girshick, K.~He, and P.~Doll{\'a}r, ``{Focal loss for
  dense object detection},'' in \emph{Proceedings of the IEEE International
  Conference on Computer Vision}, 2017, pp. 2980--2988.

\bibitem{PDV}
J.~S. Hu, T.~Kuai, and S.~L. Waslander, ``{Point density-aware voxels for LiDAR
  3D object detection},'' in \emph{CVPR}, 2022, pp. 8469--8478.

\bibitem{PV-rcnn++}
S.~Shi, L.~Jiang, J.~Deng, Z.~Wang, C.~Guo, J.~Shi, X.~Wang, and H.~Li,
  ``{PV-RCNN$++$: Point-voxel feature set abstraction with local vector
  representation for 3D object detection},'' \emph{International Journal of
  Computer Vision}, vol. 131, no.~2, pp. 531--551, 2023.

\bibitem{Voxel-rcnn}
J.~Deng, S.~Shi, P.~Li, W.~Zhou, Y.~Zhang, and H.~Li, ``{Voxel R-CNN: Towards
  high performance voxel-based 3D object detection},'' in \emph{AAAI}, vol.~35,
  no.~2, 2021, pp. 1201--1209.

\bibitem{KITTI}
A.~Geiger, P.~Lenz, and R.~Urtasun, ``{Are we ready for autonomous driving? The
  KITTI vision benchmark suite},'' in \emph{CVPR}, 2012, pp. 3354--3361.

\bibitem{OpenPCDet}
O.~D. Team, ``{OpenPCDet: An open-source toolbox for 3D object detection from
  point clouds},'' \url{https://github.com/open-mmlab/OpenPCDet}, 2020.

\bibitem{focal}
Y.~Chen, Y.~Li, X.~Zhang, J.~Sun, and J.~Jia, ``{Focal sparse convolutional
  networks for 3D object detection},'' in \emph{CVPR}, 2022.

\end{thebibliography}

\end{document}